\documentclass[lettersize,journal]{IEEEtran}
\usepackage{amsmath,amsfonts}
\usepackage{algorithmic}
\usepackage{algorithm}
\usepackage{array}
\usepackage[caption=false,font=normalsize,labelfont=sf,textfont=sf]{subfig}
\usepackage{textcomp}
\usepackage{stfloats}
\usepackage{url}
\usepackage{verbatim}
\usepackage{graphicx}
\usepackage{cite}
\usepackage{fontawesome} 
\usepackage{simpleicons} 
\usepackage{xcolor} 
\usepackage{amsmath}
\usepackage{amssymb}
\usepackage{multirow}
\usepackage{graphicx}
\usepackage{diagbox}
\usepackage{hyperref}
\usepackage{threeparttable}
\begin{document}

\title{CommonVoice-SpeechRE and RPG-MoGe: Advancing Speech Relation Extraction with a New Dataset and Multi-Order Generative Framework}

\author{%
	Jinzhong~Ning, Paerhati~Tulajiang, Yingying~Le, Yijia~Zhang, Yuanyuan~Sun, Hongfei~Lin, Haifeng~Liu%
	
	\thanks{Jinzhong Ning and Yijia Zhang are with the School of Information Science and Technology, Dalian Maritime University, Dalian 116026, China (e-mail: ningjinzhong@dlmu.edu.cn, zhangyijia@dlmu.edu.cn).}%
	
	\thanks{Paerhati Tulajiang is with the School of Computer Science and Technology, Dalian University of Technology, Dalian 116024, China; and with the College of Computer Science and Technology, Xinjiang Normal University, Urumqi 830054, China (e-mail: prht@mail.dlut.edu.cn).}

	\thanks{Haifeng Liu is with the School of Computer and Electronic Information, Nanjing Normal University, Nanjing 210046, China; and with the Adolescent Education and Intelligence Support Lab of Nanjing Normal University, Laboratory of Philosophy and Social Sciences at Universities in Jiangsu Province (e-mail: liuhaifeng@nnu.edu.cn).}

	\thanks{Yingying Le, Yuanyuan Sun, and Hongfei Lin are with the School of Computer Science and Technology, Dalian University of Technology, Dalian 116024, China (e-mail: 29354772@mail.dlut.edu.cn, syuan@dlut.edu.cn, hflin@dlut.edu.cn).}%
}

\markboth{Journal of \LaTeX\ Class Files,~Vol.~14, No.~8, August~2021}%
{Shell \MakeLowercase{\textit{et al.}}: A Sample Article Using IEEEtran.cls for IEEE Journals}


\maketitle

\begin{abstract}
Speech Relation Extraction (SpeechRE) aims to extract relation triplets directly from speech.
However, existing benchmark datasets rely heavily on synthetic data, lacking sufficient quantity and diversity of real human speech. Moreover, Existing models also suffer from rigid single-order generation templates and weak semantic alignment, substantially limiting their performance. To address these challenges, we introduce \textbf{CommonVoice-SpeechRE}, a large-scale dataset comprising nearly 20,000 real-human speech samples from diverse speakers, establishing a new benchmark for SpeechRE research. Furthermore, we propose the \textbf{R}elation \textbf{P}rompt-\textbf{G}uided \textbf{M}ulti-\textbf{O}rder \textbf{G}enerative \textbf{E}nsemble (\textbf{RPG-MoGe}), a novel framework that features: (1) a multi-order triplet generation ensemble strategy, leveraging data diversity through diverse element orders during both training and inference, and (2) CNN-based latent relation prediction heads that generate explicit relation prompts to guide cross-modal alignment and accurate triplet generation. Experiments show our approach outperforms state-of-the-art methods, providing both a benchmark dataset and an effective solution for real-world SpeechRE. The source code and dataset are publicly available at \url{https://github.com/NingJinzhong/SpeechRE_RPG_MoGe}.
\end{abstract}


\section{Introduction}

\IEEEPARstart{R}{elation} Extraction (RE), a fundamental task in information extraction, aims to extract structured knowledge in the form of relational triples (head entity, relation, tail entity) from unstructured data. RE plays a pivotal role in downstream applications such as knowledge graph construction and search engine optimization \cite{NERandREsurvery}. Despite its importance, most existing research focuses on \textbf{TextRE}, which extracts relational triples solely from plain text \cite{spanbert,tplinker,rebel}.

However, with the exponential growth of speech data from sources such as news broadcasts, online meetings, and social media, there is a pressing need to extend RE to the speech domain. Speech data contains rich structured knowledge that can enhance knowledge graphs and support speech-related applications. This has led to the emergence of \textbf{Speech Relation Extraction (SpeechRE)}, a task that directly extracts relational triples from audio recordings.

\begin{figure*}[tb]
	\centering
	\includegraphics[width=\linewidth]{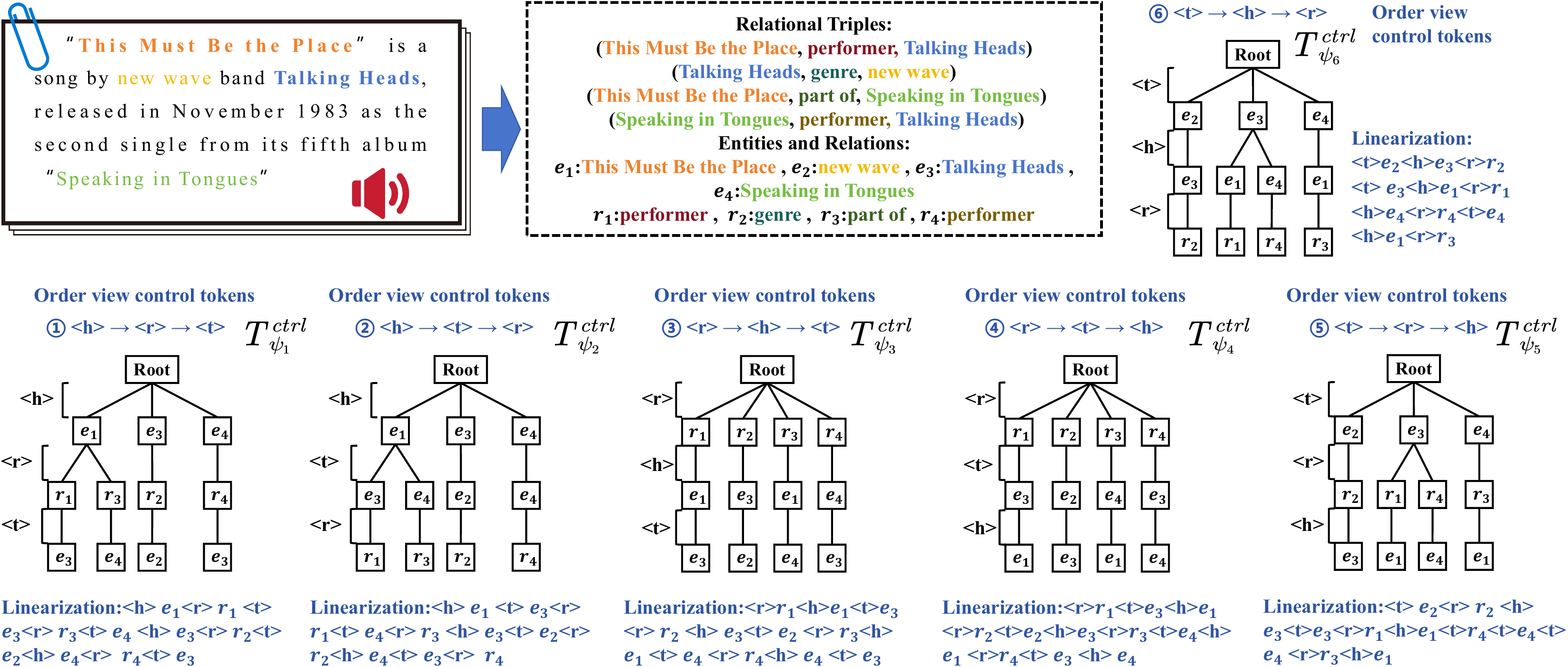}
	\caption{Explanation of the multi-view relation tree and its linearization process. Here, ``$<h>$'', ``$<r>$'', and ``$<t>$'' are special tokens representing the head entity, relation type, and tail entity of the relational triple respectively.}
	\label{multiviewretree}
\end{figure*}

Overall, SpeechRE is a relatively new research topic and remains underexplored. However, two notable works, LNA-ED \cite{LNA-ED} and MCAM \cite{MCAM}, have already made significant contributions. Wu et al. \cite{LNA-ED} introduced the SpeechRE task by applying text-to-speech (TTS) to TextRE datasets, creating two synthetic speech benchmarks. They also provided the first SpeechRE baseline, LNA-ED, which uses a CNN-based length adapter to bridge a speech encoder and text decoder. Building on this, Zhang et al. \cite{MCAM} developed two real-human-speech SpeechRE datasets and proposed MCAM, a more powerful model that employs a Multi-Level Cross-Modal Alignment Adapter to align tokens, entities, and sentences across speech and text.

Despite these advancements, existing approaches suffer from several limitations: (1) \textbf{Issue-1}: In their datasets, real-human speech data mainly covers the test set, leaving limited training examples with few speakers (see Table \ref{datasetscompare}). This may reduce the model's performance and generalization in real-world scenarios. (2) \textbf{Issue-2}: Current methods generate relational triples in a fixed order, ignoring the inherent diversity in the order of triple elements within the data. This restricts the model’s ability to fully exploit the data. (3) \textbf{Issue-3}: Existing approaches primarily rely on semantic similarity for cross-modal alignment, overlooking high-level structured semantic cues such as entity relations.

\begin{table}[tb]
	\caption{Comparison of Key Statistics between existing datasets and the dataset proposed in this paper}
	\resizebox{\linewidth}{!}
	{
	\begin{threeparttable}
		\begin{tabular}{llll}
			\hline
			\textbf{Dataset} & \textbf{CoNLL04} & \textbf{ReTACRED} & \textbf{Ours} \\ \hline
			\#Rel. & 5 & 40 & 45 \\
			\#Train Sam. & 922\textcolor{blue}{\faUser} & 33,477\textcolor{red}{\faAndroid} & 14,557\textcolor{blue}{\faUser} \\
			\#Dev Sam. & 231\textcolor{blue}{\faUser} & 9,350\textcolor{red}{\faAndroid} & 2,495\textcolor{blue}{\faUser} \\
			\#Test Sam. & 288\textcolor{blue}{\faUser} & 5,805\textcolor{blue}{\faUser} & 2,494\textcolor{blue}{\faUser} \\
			\#Speaker & 4 & 8 & $\sim$20,000 \\ \hline
		\end{tabular}
	\begin{tablenotes}
		\footnotesize
		\item \textbf{Notes}: ``\#Rel'': Number of Relations; ``Sam.'': Samples; \textcolor{blue}{\faUser}: Indicates samples with real-human speech; \textcolor{red}{\faAndroid}: Indicates samples with TTS synthetic speech
	\end{tablenotes}
	\end{threeparttable}
	}
	\label{datasetscompare}
\end{table}

To address these challenges, we propose a comprehensive solution that encompasses both data and model innovations. 

For the data limitation (\textbf{Issue-1}), we introduce CommonVoice-SpeechRE, \textbf{the first large-scale dataset with real human speech for SpeechRE}, comprising nearly 20,000 naturally spoken recordings from diverse speakers.  Our dataset establishes an authentic human speech benchmark with substantial variety of speaker profiles and scenarios (see TABLE \ref{datasetscompare}).

For the model architecture, we propose the \textbf{R}elation \textbf{P}rompt-\textbf{G}uided \textbf{M}ulti-\textbf{O}rder \textbf{G}enerative \textbf{E}nsemble (\textbf{RPG-MoGe}) framework, which systematically addresses the identified challenges through two key innovations: (1) To overcome the template rigidity in triplet generation (\textbf{Issue-2}), we introduce an innovative multi-view relation tree structure (illustrated in Figure \ref{multiviewretree}) that comprehensively captures diverse element ordering patterns. Through tree linearization as generation targets, our model implements a \textit{multi-order triplet generation ensemble strategy} across both training and inference phases, thereby maximizing the utilization of inherent data diversity. (2) To resolve the alignment deficiency (\textbf{Issue-3}), we design a \textit{CNN-based latent relation prediction head} that extracts implicit relational cues from speech signals. These are transformed into \textit{explicit relation prompts} that dynamically guide the text decoder during both triple generation and cross-modal alignment.

To our knowledge, RPG-MoGe is the first SpeechRE framework that holistically address cross-modal relation extraction challenges through three three novel components (multi-order ensemble, latent relation prediction, and explicit prompt guidance for decoder). Our contributions can be summarized as follows:
\begin{itemize}
	\item We introduce  the first large-scale, diverse real-human-speech dataset--CommonVoice-SpeechRE, establishing a new benchmark for SpeechRE research.
	\item We propose RPG-MoGe, a novel framework that integrates multi-order triple generation and explicit relation prompts to fully exploit data diversity and high-level semantic cues. 
	\item Extensive experiments on multiple SpeechRE benchmarks show that our approach outperforms state-of-the-art baselines, validating the effectiveness of our dataset and model design.
\end{itemize}

\section{Related Work}
\subsection{Speech Relation Extraction}
Relation Extraction (RE) is a fundamental task in information extraction that aims to extract structured knowledge in the form of (head entity, relation, tail entity) triplets from unstructured data. Extensive research has established three dominant paradigms for text-based RE (TextRE): (1) sequence labeling approaches \cite{spanbert,BiRTE}, (2) table-filling methods \cite{tplinker,odrte}, and (3) text generation-based frameworks \cite{rebel}. 

With the exponential growth of speech data from sources like podcasts and video content, Speech Relation Extraction (SpeechRE) has emerged as a critical yet underexplored task bridging Information Extraction (IE) and Spoken Language Understanding (SLU) \cite{sner_slue}. While substantial progress has been made in related tasks like Speech Named Entity Recognition \cite{sner_en, sner_fr, sner_cn}, SpeechRE remains nascent. Two seminal works have shaped this field: \cite{LNA-ED} pioneered the SpeechRE task by synthesizing benchmarks through TTS conversion of TextRE datasets, proposing the LNA-ED model with a CNN-based length adapter to bridge speech encoders and text decoders. Subsequently, \cite{MCAM} advanced the field by constructing the first real-human-speech dataset and developing the MCAM framework featuring hierarchical cross-modal alignment at token, entity, and sentence levels.

\subsection{Multi-view Prompt Text Generation}
Recent work in aspect-based sentiment analysis has shown that leveraging element order diversity in triples \cite{mvp} or quadruples \cite{bvsp} during training and inference can enhance model performance and generalization. Inspired by this, we are the first to explore the impact of element order diversity in relational triplets on model performance in SpeechRE, a cross-modal text generation task involving both speech and text. This approach distinguishes our work from prior research and opens new avenues for improving SpeechRE through structured data diversity.

\begin{figure*}[tb]
	\centering
	\includegraphics[width=0.9\linewidth]{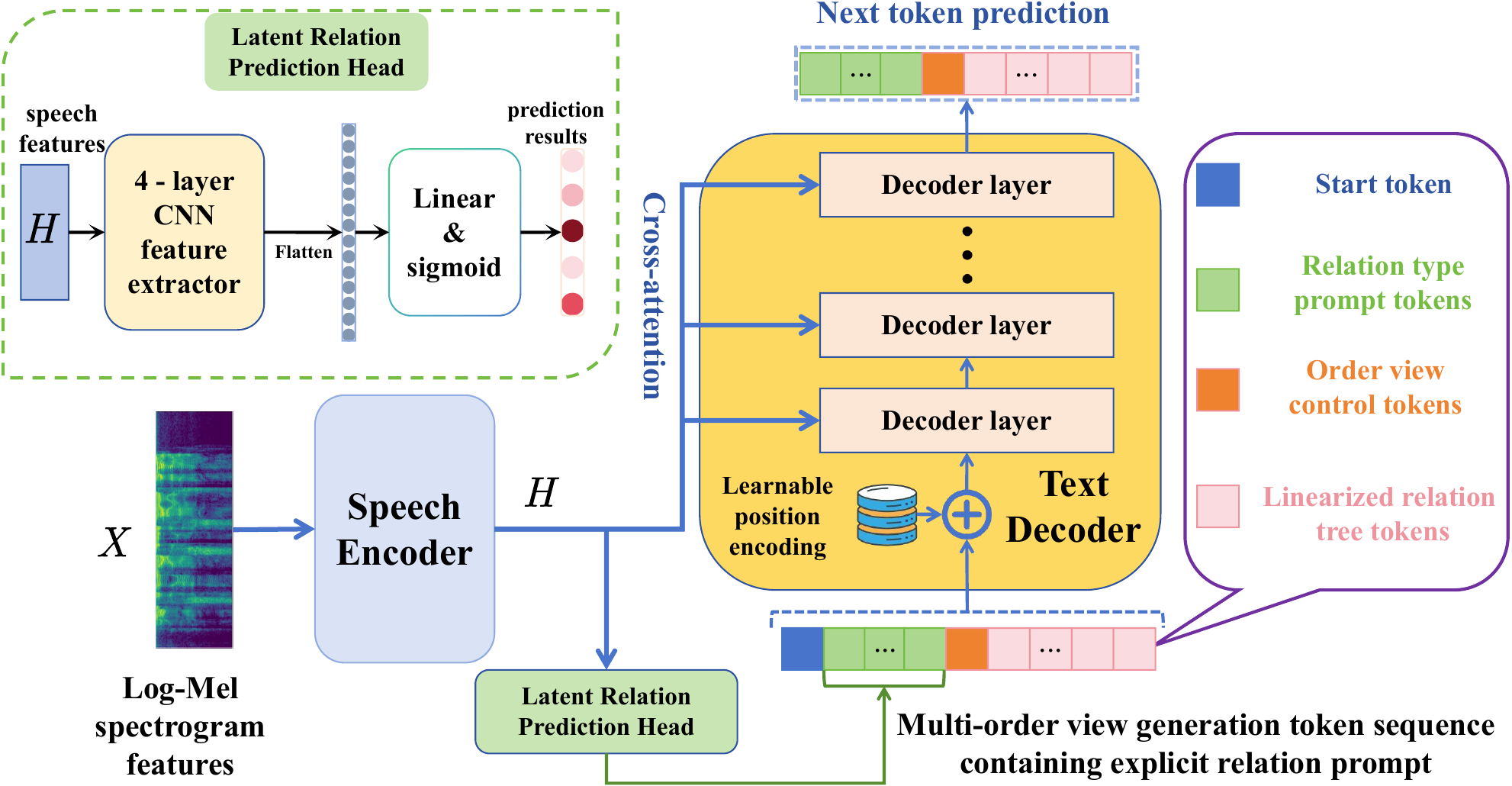}
	\caption{The overall architecture of RPG-MoGe.}
	\label{mainfigure}
\end{figure*}

\begin{figure}[tb]
	\centering
	\includegraphics[width=0.8\linewidth]{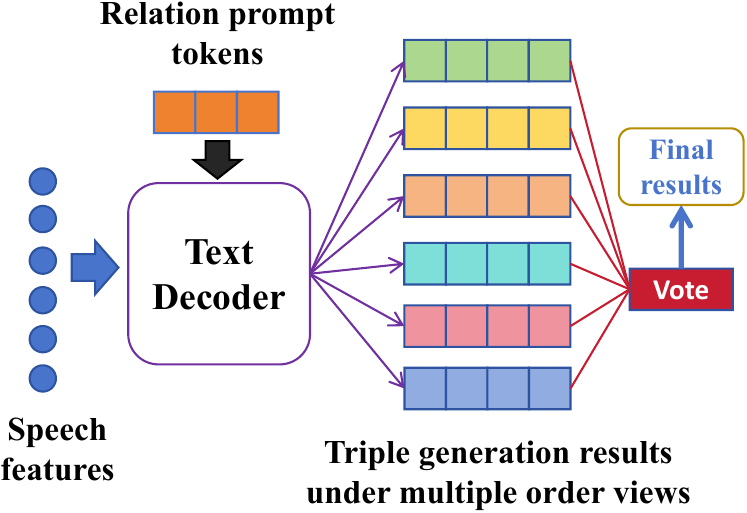}
	\caption{Implementation details for the Inference Phase in RPG-MoGe.}
	\label{RPG_infer}
\end{figure}

\section{The New Dataset}

We present CommonVoice-SpeechRE, a novel dataset derived from the English subset of the Common Voice 17.0 corpus \cite{commonvoiceASR}. Common Voice 17.0 is a large-scale, multilingual speech dataset comprising 20,408 validated hours of recordings across 124 languages, contributed by volunteers globally. Released under the CC-0 license, it permits unrestricted use, modification, and redistribution, making it an ideal foundation for secondary annotation tasks such as Speech Relation Extraction (SpeechRE).

Most samples in Common Voice 17.0 are negative examples lacking entities or relations. To identify potential positive samples, we employed a pre-trained BERT NER tagger\footnote{\href{https://huggingface.co/flair/ner-english-ontonotes}{https://huggingface.co/flair/ner-english-ontonotes}} to analyze transcriptions and filter relevant data. We adopted entity and relation type definitions from the ACE04 and ACE05 datasets, crafting a tailored annotation guide. A team of 10 graduate students (all CET-6 certified) manually labeled approximately 20,000 transcriptions using Label Studio\footnote{\href{https://labelstud.io}{https://labelstud.io}} \cite{LabelStudio}. The annotation process involved dividing the data into batches of no more than 1,000 sentences, with 10\% randomly selected for verification. Experienced annotators ensured sentence-level accuracy exceeded 95\%; otherwise, the batch was re-annotated.

\section{Methodology}

In this section, we formally define the Speech Relation Extraction (SpeechRE) task and present the detailed implementation of our proposed RPG-MoGe framework.

\subsection{Task Definition}

Given a speech signal \( \boldsymbol{S} \), the SpeechRE task aims to directly extract a set of relational triples \(\varGamma = \{ (h_i, r_i, t_i) \mid h_i, t_i \in \text{E}, r_i \in \text{R} \}\) from the speech signal, where \(\text{E}\) denotes the set of entities in the speech transcript, and \(\text{R}\) represents the set of predefined relations.

\subsection{Details of the RPG-MoGe Framework}
The ERP-MoGe framework consists of three core modules: a Speech Encoder, a Latent Relation Prediction Head, and a Text Decoder. The detailed structure is illustrated in Figure \ref{mainfigure}.
\subsubsection{Speech Encoder}

Given an input raw speech signal $\boldsymbol{S}$, we first convert it into log-mel spectrogram features $\boldsymbol{X}$. Subsequently, the features $\boldsymbol{X}$ are fed into the Whisper speech encoder \cite{whisper} to extract high-level speech features $\boldsymbol{H}$ of the speech:
\begin{equation}
	\boldsymbol{H} = \text{WhisperEncoder}(\boldsymbol{X})\in \mathbb{R} ^{L_H\times d_h}
\end{equation}
where $\text{WhisperEncoder}(\cdot)$ represents the encoding operation of the Whisper encoder model, $L_H$ and $d_h$ are sequence length and dimension of speech features $\boldsymbol{H}$.

\subsubsection{Latet Relation Prediction Head}

The Latent Relation Prediction Head (LRPH) is designed to leverage semantic entity-relation cues by predicting latent relations in the speech signal. It consists of the following steps:

CNN Layers: We pass \(\boldsymbol{H}\) through four CNN layers with ReLU activation to capture local patterns:
\begin{equation}
	\boldsymbol{H}_{\text{cnn}} = \text{Conv}_4(\boldsymbol{H})
\end{equation}

Flattening and Linear Transformation: The CNN output is flattened and fed into a linear layer to compute relation prediction scores:
\begin{gather}
	\boldsymbol{H}_{\text{flat}} = \text{Flatten}(\boldsymbol{H}_{\text{cnn}})\\
	\boldsymbol{score}^{(R)} = \sigma \left( \boldsymbol{W}_{\text{lrp}}\boldsymbol{H}_{\text{flat}}+\boldsymbol{b}_{\text{lrp}} \right)
\end{gather}
where \(\sigma\) is the sigmoid function, \(\boldsymbol{W}_{\text{lrp}} \in \mathbb{R}^{|\text{R}| \times d_h}\) and \(\boldsymbol{b}_{\text{lrp}} \in \mathbb{R}^{|\text{R}|}\) are learnable parameters, and \(\boldsymbol{score}^{(R)} \in \mathbb{R}^{|\text{R}|}\) represents the scores for all predefined relation types.

Loss Function: We employ the Binary Cross Entropy (BCE) loss for training the LRPH module:
\begin{equation}
	\begin{split}
		\mathcal{L}_{\text{lrp}} =& -\frac{1}{|\text{R}|} \sum_{i = 1}^{|\text{R}|} \left[ \text{y}_i^{(R)} \log(\text{score}_i^{(R)}) \right. \\
		& \left. + (1 - \text{y}_i^{(R)}) \log(1 - \text{score}_i^{(R)}) \right]
	\end{split}
\end{equation}
where \(\text{y}^{(R)}\) denotes the ground-truth relation labels. Since each sample may contain multiple relations, this prediction task is a multi-label classification problem. In \(\text{y}^{(R)}\), each element \(\text{y}_i^{(R)}\) can be either 0 or 1, indicating the absence or presence of the \(i\)-th relation type, respectively. This approach enables the model to predict multiple relations simultaneously for each given input.

\begin{table*}[tb]
	\caption{Dataset statistics.}
	\resizebox{\linewidth}{!}
	{
		\begin{threeparttable}
		\begin{tabular}{lcccccccc}
			\hline
			\multicolumn{1}{c}{\multirow{2}{*}{\textbf{Datasets}}} & \multirow{2}{*}{\textbf{\#Relations}} & \multicolumn{3}{c}{\textbf{\#Instances}}      & \multicolumn{3}{c}{\textbf{\#Triplets}}       & \multirow{2}{*}{\textbf{\#Avg. audio length}} \\
			\multicolumn{1}{c}{}                                   &                                       & \textbf{train} & \textbf{dev} & \textbf{test} & \textbf{train} & \textbf{dev} & \textbf{test} &                                               \\ \hline
			\textcolor{red}{\faAndroid}CoNLL04-SpeechRE                                       & 5                                     & 922            & 231          & 288           & 1,283          & 343          & 422           & 11.3s                                          \\
			\textcolor{red}{\faAndroid}ReTACRED-SpeechRE                                      & 40                                    & 33,477         & 9,350        & 5,805         & 58,465         & 19,584       & 13,418        & 12.9s                                          \\
			\textcolor{blue}{\faUser}CommonVoice-SpeechRE                                   & 45                                    & 14,557         & 2,495        & 2,494         &15,948                &2,696              &2,728               & 11.6s                                          \\ \hline
		\end{tabular}%
		\begin{tablenotes}
			\footnotesize
			\item \textbf{Notes}: \textcolor{red}{\faAndroid}: TTS-synthesized speech; \textcolor{blue}{\faUser}: real human speech. ReTACRED-SpeechRE enumerates all entity pairs as triplets, including ``no\_relation'' type, while the other two datasets only contain positive triplets.
		\end{tablenotes}
		\end{threeparttable}
	}
	\label{datasetsstatistics}
\end{table*}

\subsubsection{Multi-view Relation Tree and Linearization}

To model the diversity introduced by permutations of triplet element orders, we propose the Multi-view Relation Tree structure. As depicted in Figure \ref{multiviewretree}, each tree consists of four layers, with each layer (excluding the first) corresponding to an element of the triple. For a given sample, we can generate \(P(3,3) = 6\) distinct relation trees by permuting the order of triplet elements.

Formally, for a speech signal \(\boldsymbol{S}\) with a set of relation triplets \(\mathcal{T}\), we apply the \(\text{Treeify}(\cdot,\cdot)\) function to construct a relation tree \(\mathcal{G}_{\psi_i}\) from a specific order perspective \(\psi_i\):
\begin{equation}
	\mathcal{G}_{\psi_i} = \text{Treeify}(\mathcal{T}, \psi_i)
\end{equation}
where \(\psi_i \in \varPsi\) represents an order perspective, and \(\varPsi\) encompasses all six possible order perspectives.

The relation tree \(\mathcal{G}_{\psi_i}\) is then linearized into a token sequence using the \(\text{SeqLin}(\cdot)\) operation:
\begin{equation}
	T_{lin}^{\psi_i} = \text{SeqLin}(\mathcal{G}_{\psi_i})
\end{equation}

\subsubsection{Text Decoder}   

The Text Decoder uses relation prompts and multi-order triplet generation to decode relational triplets. We utilize the pre-trained Whisper decoder \cite{whisper} for this purpose. The input token sequence to the decoder consists of three parts:

Relation type prompt tokens: \(\boldsymbol{T}_{rel} = [t_1^{rel}, \ldots, t_n^{rel}]\), where \(t_i^{rel}\) are special tokens representing the predicted relation types generated by the Latent Relation Prediction Head. These tokens guide the decoder by incorporating latent relational cues from speech.

Order view control tokens: \(\boldsymbol{T}_{\psi_i}^{ctrl} = \text{permute}\left( \left[ \left< h \right>, \left< r \right>, \left< t \right> \right], \psi_i \right)\), which specify the order of special tokens $\left< h \right> ,\left< r \right> ,\left< t \right>$ for a given perspective \(\psi_i\), as illustrated in Figure \ref{multiviewretree}.

Linearized relation tree tokens: \(\boldsymbol{T}_{\psi_i}^{lin}\), which represent the linearized token sequence of the relation tree. This component encodes the hierarchical structure of the relation tree into a sequential format suitable for the decoder.

These components are concatenated into the decoder input sequence \(\boldsymbol{T}_{dec} = [\boldsymbol{T}_{rel}, \boldsymbol{T}_{\psi_i}^{ctrl}, \boldsymbol{T}_{\psi_i}^{lin}]\). At the \(i\)-th decoding step, the probability distribution \(\boldsymbol{p}_{t_{i}^{dec}}\) of the output token \(t_{i}^{dec}\) is computed as:
\begin{gather}
	\boldsymbol{h}_{t_{i}^{dec}}=\text{WhisperDecoder}\left( \boldsymbol{H},\boldsymbol{T}_{dec}^{<i} \right)\\
	\boldsymbol{p}_{t_{i}^{dec}}=\text{Softmax} \left( \boldsymbol{W}_{lm}\boldsymbol{h}_{t_{i}^{dec}}+\boldsymbol{b}_{lm} \right)
\end{gather}
where \(\boldsymbol{h}_{t_{i}^{dec}}\) is the hidden state, and \(\boldsymbol{W}_{lm}\), \(\boldsymbol{b}_{lm}\) are learnable parameters.

The decoder is trained using the Cross-Entropy Loss:
\begin{equation}
	\mathcal{L}_{dec}=-\frac{1}{N}\sum_{i = 1}^{N}\sum_{j = 1}^{|V|}\boldsymbol{y}_{t_{i}^{dec}}[j]\log(\boldsymbol{p}_{t_{i}^{dec}}[j])
\end{equation}
where \(N\) is the sequence length, \(|V|\) is the vocabulary size, and \(\boldsymbol{y}_{t_{i}^{dec}}\) is the token label at the $i$-th decoding step.

\subsubsection{Training and Inference Strategies}
\label{TrainingandInferenceStrategies}

\textbf{During training}, each sample is expanded into multiple generation targets corresponding to all possible order views for participation in training. The total loss combines the $\mathcal{L}_{\text{lrp}}$ and $\mathcal{L}_{dec}$:
\begin{equation}
	\mathcal{L}_{total} = \mathcal{L}_{\text{lrp}} + \mathcal{L}_{dec}
\end{equation} 

To address potential discrepancies between training and testing performance in the Latent Relation Prediction Head (LRPH), we implemented two key regularization techniques: (1) Incorporated dropout layers (p=0.5) within the LRPH module, and (2) Adopted a training strategy that randomly masks 0-50\% of positive relations predicted by LRPH during training (disabled during inference). 

\textbf{During inference}, as illustrated in Figure \ref{RPG_infer}, the text decoder takes the speech features \(\boldsymbol{H}\) and relation prompt tokens \(\boldsymbol{T}_{rel}\) as initial inputs. By varying the order view control tokens, the decoder autoregressively generates triplets under all order views. A triplet is included in the final results if it appears in more than \(\lambda_{vote}\) order views.

\begin{table*}[ht]
	\caption{F1-score (\%) comparison: RPG-MoGe versus baselines.}
	\resizebox{\textwidth}{!}{%
		\begin{threeparttable}
		\begin{tabular}{llcccccccccc}
			\hline
			\multicolumn{2}{c}{\multirow{2}{*}{\textbf{Model}}}                                                                              & \multirow{2}{*}{\textbf{\begin{tabular}[c]{@{}c@{}}External\\ Resources\end{tabular}}} & \multicolumn{3}{c}{\textcolor{red}{\faAndroid}\textbf{CoNLL04-SpeechRE}}          & \multicolumn{3}{c}{\textcolor{red}{\faAndroid}\textbf{ReTACRED-SpeechRE}}         & \multicolumn{3}{c}{\textcolor{blue}{\faUser}\textbf{CommonVoice-SpeechRE}}      \\ \cline{4-12} 
			\multicolumn{2}{c}{}                                                                                                             &                                                                                        & \textbf{Entity} & \textbf{Relation} & \textbf{Triplet} & \textbf{Entity} & \textbf{Relation} & \textbf{Triplet} & \textbf{Entity} & \textbf{Relation} & \textbf{Triplet} \\ \hline
			\multicolumn{1}{c}{\multirow{7}{*}{TextRE}}                                   & GPT-3.5(LLM)                                     & -                                                                                      & 58.74           & 49.45             & 22.27            & 40.46           & 17.63             & 3.22             & 53.74           & 28.41             & 10.73            \\
			\multicolumn{1}{c}{}                                                          & GPT-4(LLM)                                       & -                                                                                      & 61.36           & 62.67             & 28.83            & 47.4            & 39.12             & 9.12             & 57.33           & 38.32             & 15.35            \\
			\multicolumn{1}{c}{}                                                          & TP-Linker                                        & -                                                                                      & 78.63           & 83.49             & 58.56            & 50.46           & 51.83             & 20.39            & 64.61           & 69.31             & 46.61            \\
			\multicolumn{1}{c}{}                                                          & Spert                                            & -                                                                                      & 76.38           & 81.83             & 63.45            & 60.26           & 63.48             & 21.46            & 66.34           & 70.82             & 47.26            \\
			\multicolumn{1}{c}{}                                                          & REBEL                                            & -                                                                                      & 85.36           & 89.86             & 71.46            & 60.09           & 65.15             & 25.15            & 71.32           & 74.32             & 49.81            \\
			\multicolumn{1}{c}{}                                                          & BiRTE                                            & -                                                                                      & 79.34           & 87.34             & 64.61            & 61.93           & 65.51             & 20.76            & 67.61           & 72.35             & 47.10            \\
			\multicolumn{1}{c}{}                                                          & OD-RTE                                           & -                                                                                      & 81.81           & 82.35             & 60.57            & 59.37           & 60.94             & 21.08            & 70.61           & 69.67             & 47.34            \\ \hline
			\multirow{7}{*}{\begin{tabular}[c]{@{}l@{}}SpeechRE\\ (Pipline)\end{tabular}} & GPT-3.5$_{pipe}$(LLM)                            & -                                                                                      & 28.21           & 69.61             & 6.31             & 16.61           & 43.84             & 1.32             & 21.30           & 46.81             & 3.34             \\
			& GPT-4$_{pipe}$(LLM)                              & -                                                                                      & 29.41           & 70.31             & 7.13             & 19.76           & 46.31             & 4.23             & 23.61           & 44.35             & 4.94             \\
			& TP-Linker$_{pipe}$                               & -                                                                                      & 35.21           & 78.21             & 9.76             & 30.27           & 50.01             & 6.59             & 31.06           & 64.13             & 7.61             \\
			& Spert$_{pipe}$                                   & -                                                                                      & 30.43           & 75.95             & 11.88            & 34.36           & 57.17             & 6.89             & 32.61           & 64.48             & 7.54             \\
			& REBEL$_{pipe}$                                   & -                                                                                      & 37.06           & 83.35             & 14.01            & 32.07           & 51.97             & 6.49             & 31.54           & 66.10             & 7.92             \\
			& BiRTE$_{pipe}$                                   & -                                                                                      & 31.94           & 76.61             & 12.34            & 32.55           & 56.63             & 6.73             & 32.56           & 65.12             & 7.64             \\
			& OD-RTE$_{pipe}$                                  & -                                                                                      & 34.36           & 79.23             & 9.83             & 31.34           & 51.32             & 6.64             & 31.67           & 65.15             & 7.56             \\ \hline
			\multirow{8}{*}{\begin{tabular}[c]{@{}l@{}}SpeechRE\\ (End2End)\end{tabular}} & GPT-4o-audio(LLM)                                & -                                                                                      & 31.21           & 59.57             & 5.64             & 13.21           & 41.61             & 1.14             & 29.33           & 31.70             & 3.12             \\
			& Qwen2-audio(LLM)                                 & -                                                                                      & 36.74           & 16.31             & 2.31             & 10.50           & 23.61             & 0.31             & 31.16           & 14.92             & 0.85             \\
			& LNA-ED(520M)$_{ori}$                             & PL-FT                                                                                  & 18.87           & 55.66             & 10.41            & 17.21           & 43.37             & 3.20             & 26.34           & 37.31             & 5.37             \\
			& LNA-ED(770M)$_{whi}$                             & -                                                                                      & 19.13           & 56.32             & 11.12            & 18.26           & 43.15             & 3.67             & 27.61           & 38.51             & 6.01             \\
			& MCAM(520M)$_{ori}$                               & ASR-PTC                                                                                & 40.13           & 77.89             & 22.07            & 35.34           & 58.96             & 8.07             & 43.94           & 48.37             & 14.96            \\
			& MCAM(770M)$_{whi}$                               & -                                                                                      & 40.66           & 77.61             & 22.71            & 35.61           & \textbf{59.13}    & 8.21             & 45.34           & 50.34             & 15.71            \\
			& \textbf{RPG-MoGe}(250M)$_{whi}$ & -                                                                                      & 43.16           & 76.91             & 22.17            & 36.00           & 57.46             & 8.09             & 45.59           & 49.60             & 15.32            \\
			& \textbf{RPG-MoGe}(770M)$_{whi}$ & -                                                                                      & \textbf{50.21}  & \textbf{79.64}    & \textbf{24.67}   & \textbf{36.76}  & 58.38             & \textbf{9.18}    & \textbf{47.20}  & \textbf{53.48}    & \textbf{18.29}   \\ \hline
		\end{tabular}%
		\begin{tablenotes}
			\footnotesize
			\item \textbf{Notes}: Subscript $pipe$ denotes ASR+TextRE pipeline methods; `PL-FT' indicates fine-tuning with pseudo-labeled data; `ASR-PTC' refers to pre-training with ASR data. Subscript $ori$ represents the original LNA-ED\cite{LNA-ED}/MCAM\cite{MCAM} backbone: 24-layer Wave2vec encoder + 12-layer BART-large decoder (520M). Subscript $whi$ denotes Whisper \cite{whisper} backbones: 24-layer encoder/decoder (770M) or 12-layer encoder/decoder (250M).
		\end{tablenotes}
		\end{threeparttable}
	}
	\label{MainResults}
\end{table*}

\section{Experiments}

\subsection{Datasets \& Evaluation Metrics}

We conducted experiments on three datasets: CoNLL04-SpeechRE, ReTACRED-SpeechRE and the CommonVoice-SpeechRE dataset proposed in this paper. The CommonVoice-SpeechRE dataset includes diverse real human speech in its training, development, and test sets. For CoNLL04-SpeechRE and ReTACRED-SpeechRE, since the real human speech test set and partial real human speech training set proposed by \cite{MCAM} have not yet been released, we used the fully TTS-generated speech version released by \cite{LNA-ED}. Detailed statistics of the datasets are provided in Table \ref{datasetsstatistics}. For evaluation metrics, following previous work \cite{LNA-ED, MCAM}, we used the micro-F1 score to assess the performance of models in entity recognition, relation prediction, and relation triplet extraction. For an entity, relation or triple to be considered correct, it must exactly match its counterpart in the ground truth tags.

\subsection{Experimental Settings}  
Our model was implemented using PyTorch-Lightning\footnote{\href{https://github.com/Lightning-AI/pytorch-lightning}{https://github.com/Lightning-AI/pytorch-lightning}} and PyTorch \cite{pytorch}, with OpenAI's Whisper\footnote{Whisper has become a standard backbone in speech processing, similar to BERT and BART in NLP.} \cite{whisper} as the backbone, specifically the whisper-small\footnote{\href{https://huggingface.co/openai/whisper-small.en}{https://huggingface.co/openai/whisper-small.en}} (244M) and whisper-medium\footnote{\href{https://huggingface.co/openai/whisper-medium.en}{https://huggingface.co/openai/whisper-medium.en}} (769M) versions. We optimized the model parameters using the Adam optimizer with a learning rate of 1e-5, a batch size of 12. Training epochs were set to 50 for CoNLL04-SpeechRE, 20 for ReTACRED-SpeechRE, and 10 for CommonVoice-SpeechRE. For the relation prediction head, we employed a four-layer CNN with 2D convolutions (kernel size = 3) and progressively increasing channel dimensions (16, 32, 64, 128). During inference, the voting threshold \(\lambda_{vote}\) for all order views was set to 2. All hyperparameters were tuned on the development set, and the best-performing checkpoint was selected for test set evaluation. Training was conducted on a single NVIDIA A40 GPU, while inference was performed on a single NVIDIA GeForce RTX 4090 GPU.

\subsection{Baselines}  
To comprehensively evaluate the performance of our proposed model, we compare it with three categories of competitive baselines: (1) \textbf{TextRE Models.} These models are designed to jointly extract entities and relations from input text. For a fair comparison, following prior works \cite{LNA-ED,MCAM}, we adopt three strong TextRE models: TP-Linker \cite{tplinker}, Spert \cite{spanbert}, and REBEL \cite{rebel} augmented with two recent advances (BiRTE \cite{BiRTE} and OD-RTE \cite{odrte}) Additionally, to explore the potential of large language models (LLMs) in relation extraction, we include GPT-3.5\footnote{gpt-3.5-turbo-0125} and GPT-4\footnote{gpt-4-turbo-2024-04-09} as baselines, leveraging their in-context learning capabilities for TextRE tasks. (2) \textbf{Pipeline SpeechRE Models.} These models follow a two-stage pipeline: first, an Automatic Speech Recognition (ASR) module transcribes the input speech into text; second, a TextRE module extracts relation triplets from the transcribed text. To ensure a fair comparison, we follow the setup of prior works \cite{LNA-ED,MCAM} and employ the pre-trained wav2vec-large model as the ASR module. For the TextRE module, we use the same five TextRE models mentioned above, resulting in seven pipeline models: TP-Linker$_{pipe}$, Spert$_{pipe}$, REBEL$_{pipe}$, BiRTE$_{pipe}$, OD-RTE$_{pipe}$, GPT-3.5$_{pipe}$, and GPT-4$_{pipe}$. (3) \textbf{End-to-End SpeechRE Models.} These models are designed to directly extract relation triplets from input speech, without the intermediate step of text transcription. Our proposed RPG-MoGe also falls into this category. As baselines, we include two existing end-to-end SpeechRE models: LNA-ED \cite{LNA-ED} and MCAM \cite{MCAM}. Additionally, to explore the capabilities of recent advancements in speech-based LLMs, we introduce two in-context learning baselines: GPT-4o-audio\footnote{\href{https://platform.openai.com/docs/models\#gpt-4o-audio}{gpt-4o-audio-preview-2024-12-17}} and Qwen2-audio\footnote{\href{https://huggingface.co/Qwen/Qwen2-Audio-7B-Instruct}{Qwen2-Audio-7B-Instruct}} \cite{Qwen2-Audio}.

\subsection{Main Results}

We conducted a comprehensive performance comparison between our proposed RPG-MoGe model and several strong baselines, including TextRE, SpeechRE (Pipeline), and SpeechRE (End2End). The experimental results, presented in Table \ref{MainResults}, reveal the following key observations:

(1) RPG-MoGe outperforms all SpeechRE (End2End) baselines, achieving state-of-the-art performance in entity, relation, and triplet F1 scores across all datasets. Notably, RPG-MoGe with a 250M parameter Whisper backbone surpasses the SOTA baseline MCAM using a 520M backbone and matches MCAM’s performance with a 770M backbone. This demonstrates RPG-MoGe’s ability to leverage the diversity of relation triplet element orders and effectively utilize high-level semantic cues through its potential relation prediction head and explicit relation prompts.

(2) RPG-MoGe consistently outperforms all SpeechRE models in triplet extraction, highlighting the limitations of the pipeline approach, where cascading ASR with TextRE introduces significant errors. The end-to-end approach effectively mitigates error accumulation, improving entity, relation, and triplet extraction accuracy.

(3) Large language models without fine-tuning (e.g., GPT-3.5, GPT-4, GPT-4o-audio, Qwen2-audio) perform significantly worse on the datasets compared to fine-tuned smaller models, emphasizing the continued importance of developing fine-tuned models in TextRE and SpeechRE domains.

(4) Replacing the non-aligned Wave2vec and BART encoders in LNA-ED and MCAM with the pre-trained and aligned Whisper encoder and decoder eliminates the need for extensive external corpus alignment and improves performance. This also ensures a fairer comparison with RPG-MoGe, which utilizes Whisper as its backbone.

\section{Conclusion}
In this work, we address the limitations of existing datasets and models in Speech Relation Extraction (SpeechRE) by introducing CommonVoice-SpeechRE, a large-scale dataset with diverse real-human speech samples, and proposing RPG-MoGe, a novel framework that leverages a multi-order triplet generation ensemble strategy and CNN-based latent relation prediction heads to enhance triple generation and cross-modal alignment. Extensive experiments demonstrate the superiority of our approach, outperforming state-of-the-art baselines and setting a new benchmark for SpeechRE research. Our contributions provide both a valuable resource and an effective methodology, advancing the field toward real-world applications.

\bibliography{main}
\bibliographystyle{IEEEtran}

\end{document}